\documentclass{article}
\usepackage{arxiv}
\usepackage{makecell}

\usepackage{pgfplots}

\usepackage{subfigure}

\usepackage{multirow}

\usepackage{hyperref}
\hypersetup{
	colorlinks=true,
	linkcolor=blue,
	filecolor=magenta,      
	urlcolor=cyan,
}

\newcommand{\mn}{MobileNet-V2}
\newcommand{\rn}{ResNet-50}
\newcommand{\nx}{Nvidia Xavier}

\usepackage{amsmath}
\newcommand{\norm}[1]{\left\lVert#1\right\rVert}

\begin{document}
\title{Real-time Person Re-identification at the Edge: A Mixed Precision Approach}

%
%
%



\author{Mohammadreza~Baharani\thanks{This is a pre-print of an article published in International Conference on Image Analysis and Recognition (ICIAR 2019), Lecture Notes in Computer Science. The final authenticated version is available online at: \href{https://doi.org/10.1007/978-3-030-27272-2_3}{https://doi.org/10.1007/978-3-030-27272-2\_3}}\\
	Electrical and Computer Engineering Department\\
	The University of North Carolina at Charlotte \\
	Charlotte, NC, 28223 USA\\
	\texttt{mbaharan@uncc.edu}\\
	\And
	Shrey~Mohan\\
	Electrical and Computer Engineering Department\\
	The University of North Carolina at Charlotte \\
	Charlotte, NC, 28223 USA\\
	\texttt{smohan7@uncc.edu}\\
	\And
	Hamed~Tabkhi\\
	Electrical and Computer Engineering Department\\
	The University of North Carolina at Charlotte \\
	Charlotte, NC, 28223 USA\\
	\texttt{htabkhiv@uncc.edu}
}

%
%
%
\maketitle              

\begin{abstract}
A critical part of multi-person multi-camera tracking is person re-identification (re-ID) algorithm, which recognizes and retains identities of all detected unknown people throughout the video stream. Many re-ID algorithms today exemplify state of the art results, but not much work has been done to explore the deployment of such algorithms for computation and power constrained real-time scenarios. In this paper, we study the effect of using a light-weight model, MobileNet-v2 for re-ID and investigate the impact of single (FP32) precision versus half (FP16) precision for training on the server and inference on the edge nodes. We further compare the results with the baseline model which uses ResNet-50 on state of the art benchmarks including CUHK03, Market-1501, and Duke-MTMC. The MobileNet-V2 mixed precision training method can improve both inference throughput on the edge node, and training time on server $3.25\times$ reaching to 27.77fps and $1.75\times$, respectively and decreases power consumption on the edge node by $1.45\times$, while it deteriorates accuracy only 5.6\% in respect to ResNet-50 single precision on the average for three different datasets. The code and pre-trained networks are publicly available
\footnote{\href{https://github.com/TeCSAR-UNCC/person-reid}{https://github.com/TeCSAR-UNCC/person-reid}}.

\keywords{Person re-identification  \and \mn~\and \rn~\and Real-time multi-target multi-camera tracking\and Edge node \and Triplet-loss}
\end{abstract}

\section{Introduction}
Real-time Multi-target Multi-Camera Tracking (MTMCT) is a task of positioning different unknown people on different camera views at a constrained amount of time. The results of this task can be beneficial to video surveillance, smart stores, and behavior analysis and anomaly detection. The core of MTMCT is person re-identification (re--ID) algorithm which retrieves person identities regardless of their poses and camera views. 

Ideally, system re-identification should happen in real-time fashion at edge nodes. However, the most of recently proposed methods in literature \cite{zhang2017alignedreid,sun2018beyond,Horizontal,ristani2018features} used \rn~\cite{ResNet} as a backbone of their method, which is computationally expensive and targeted at the server side. One approach to reducing the computation complexity is to leverage light-weight network models such as \mn~\cite{mobileNetV2}, even though these networks might not meet the real-time demands due to limited hardware resources and memory bandwidth at the edge node. Lowering accuracy and quantization can relive the pressure on both memory bandwidth and computation throughput \cite{baharani2014high,micikevicius2018mixed}. However, to the best of our knowledge, the effect of quantization on final accuracy has not been studied for object re-identification approaches based on deep learning paradigms.

In this paper, we proposed a scalable and light-weight architecture based on the \mn~to meet a predefined timing and power budget. We have also studied the effect of quantization and mixed training on two \rn~and \mn~network architecture in details. Specifically, our contribution is summarized as follows:

\begin{itemize}
   \item We proposed a re-identification architecture based on light-weight \mn, and we compare the results against the \rn~in respect of accuracy.
   \item We also studied the effect of mixed precision training approach on both \rn~and \mn. We investigated which layers of networks should be quantized to be able to train the network. Our finding is orthogonal to other object re-identification based on deep learning methods and can be applied to improve their performance.
   \item We evaluated final system performance concerning throughput and power consumption on \nx~board and discussed in details.
\end{itemize}

The rest of this article is organized as the following: Section \ref{sec:relatedWork} briefly reviews the previous person re--ID approaches. Section \ref{sec:methos} presents our re--ID methods based on MobileNet-V2 and mixed precision for real-time inference. Section \ref{sec:ExpResults} presents the experimental results including comparison with existing approaches, and finally Section \ref{sec:Conclusion} concludes this article.

\section{Related Works}\label{sec:relatedWork}
There has been an increasing amount of research in the domain of object detection and tracking in recent years. With the problem of multi-object detection and tracking comes another one, which is re-identifying the same objects throughout the frames in the video precisely as the accuracy of tracking highly depends on it. Hence, in this section, we will be reviewing some of the recent work done for person re-id.

Classical computer vision approaches like those in \cite{Tern2012CovarianceDM} which are based on covariance descriptors which augment various feature representations of an image like RGB, Hue-Saturation-Value (HSV), local binary patterns, etc. over a mean Riemannian matrix (introduced by Bak et al.\cite{Bak}) from multi-shot images to find similarities between different images have been done. Similar approaches were adopted in \cite{Mier} for real-time embedded computation but the authors do not provide with any accuracy evaluation.

In \cite{Oliveira}, Oliveira et al. generate a unique signature for each object which comprises interest points and color features for the object and calculates the similarity between different signatures using Sum of Quadratic Differences(SQD). Similar classical approaches are demonstrated by \cite{Martin} and \cite{Fleuret:192416}, which uses Biologically Inspired Features (BIF) and k-shortest path algorithm. Classical techniques are promising; however, with the boom in deep learning algorithms and plethora of computational power all thanks to the top of the line GPUs, they are even surpassing human level recognition for re-id.

Modern deep learning techniques like Alignedreid \cite{zhang2017alignedreid}, extract features from ROI using CNNs as base networks and then divide the feature map into local and global features intuitively dividing the ROI into horizontal sections and matching each section with the other images. Xiaoke et al. in \cite{Zhu}, use videos instead of separate frames to learn the inter-video and intra-video distances between people in them effectively creating triplet pairs. Tong et al. \cite{Xiao} proposed an Online Instance Matching(OIM) loss paradigm which uses a Look Up Table(LUT) for labeled objects and a cicular queue for unlabelled objects and learns to re-id people on the go. 

Yantao et al. in \cite{Shen_2018_ECCV} formulate the problem of person re-id into a graph neural network problem, with each node denoting a pair of images whose similarity and dissimilarities are learned through a  message passing technique between the nodes. Siamese network is used to compute similarity metric between pairs. Authors in \cite{Li} introduce spatiotemporal attention models to learn key spatial features of objects throughout the video. 


Almost all the works as mentioned above are novel and state of the art, however, they all use very complicated and deep networks which would hinder their performance in real-time scenarios. In next section, we propose a light-weight system with reasonably high accuracy on the state of the art benchmarks.

\section{Mixed Precision Real-time Person Re-identification}\label{sec:methos}
In this section, we discuss two \rn~and \mn~ architectures. Then we continue to explain the loss function, and in the last, we give an introduction about mixed precision training, and we elaborate more on network layers and loss precision partitioning.
\subsection{Background}
In this section, we will give an introduction for two \rn, \mn~networks. Since the \rn~ is massively used for state-of-art person re-id, we concider it as baseline for our evaluation in experimental results (Section \ref{sec:ExpResults}).
\subsubsection{Residual Network (ResNet)}
Observing the difficulty of optimizing deeper convolutional networks for the task of image recognition and image localization, authors in \cite{ResNet} came up with the idea of using shortcut connections which they called residual connections claiming that such connections will help deeper stacked networks to learn efficiently. They use the baseline VGG nets \cite{Simonyan15} architecture, which uses 3x3 convolution blocks, and translate it to a 34 layer plain network and a 34 layer residual network (ResNet) for initial experimentation. They further evaluate the results on deeper ResNets with 50, 101 and 152 layers with the latter achieving a minimum error rate on the ImageNet \cite{imagenet} dataset for recognition. 

\begin{figure*}[t]
	\centering
	
	\subfigure[Basic building block]{%
		{\includegraphics[width=0.2\textwidth, keepaspectratio, trim= 5 10 1 5,clip]{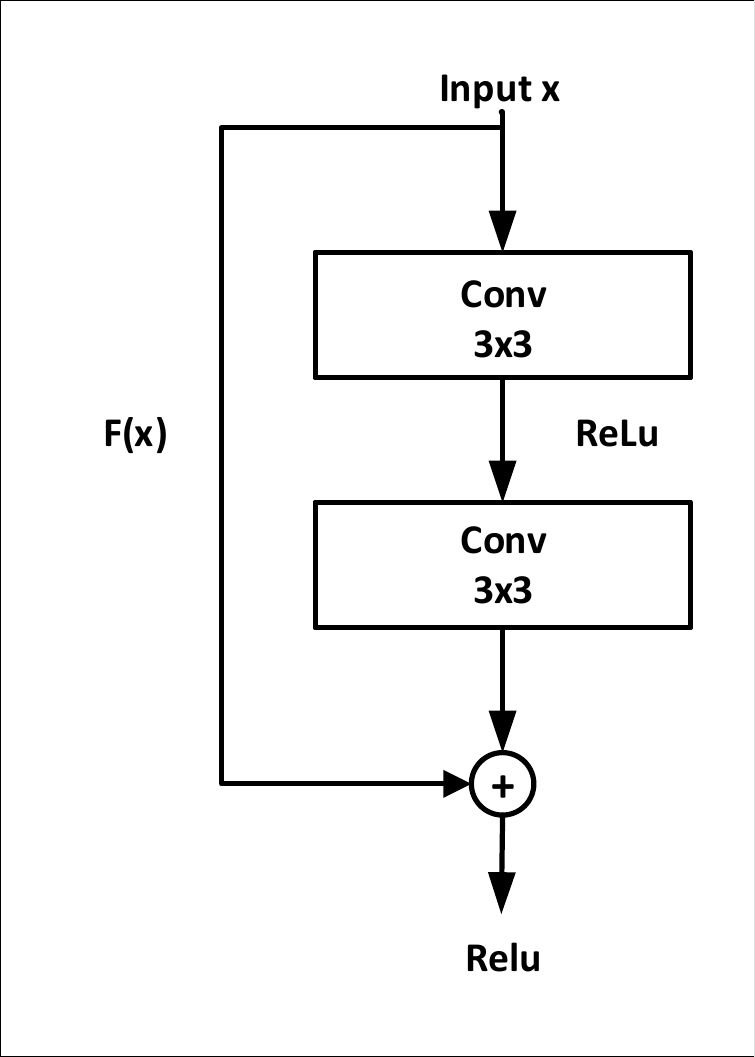}}%
		\label{fig:basicBlock}%
	}\qquad \hspace{1.5cm}
	\subfigure[Bottleneck block]{%
		{\includegraphics[width=0.15\textwidth, keepaspectratio, trim= 5 10 10 5,clip]{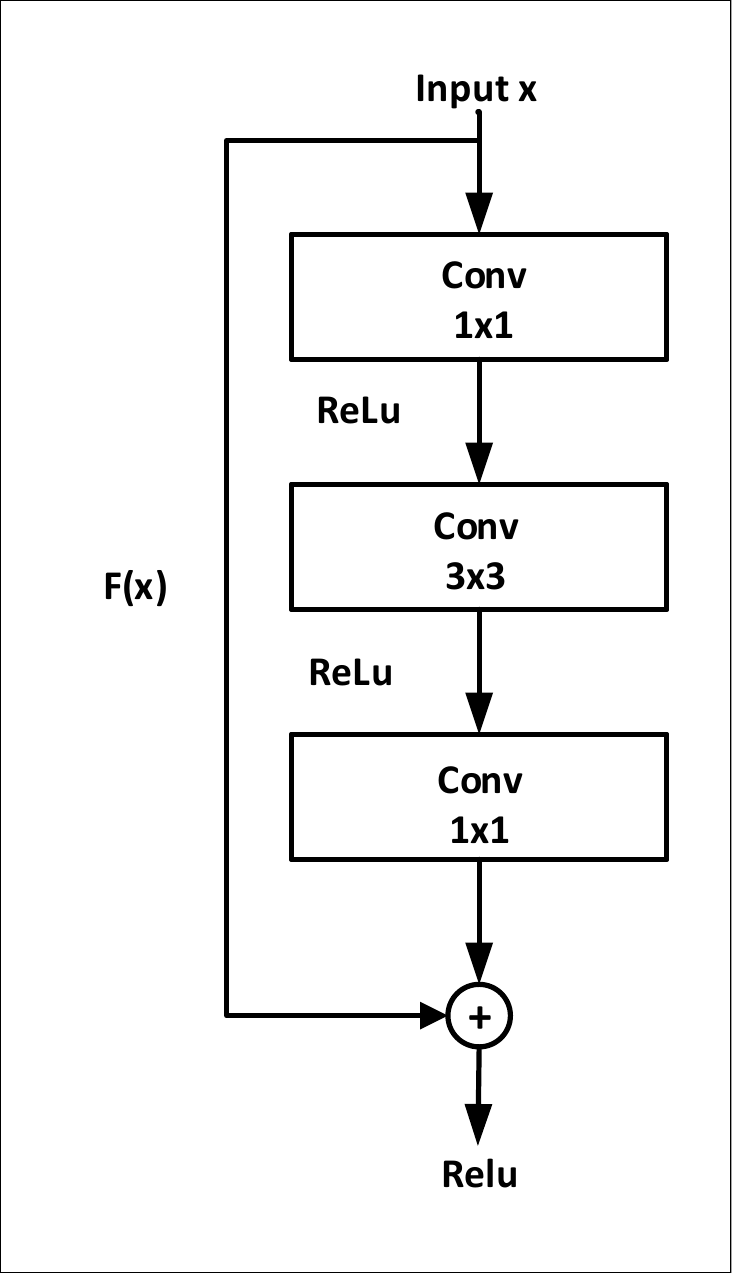}}%
		\label{fig:bottleNeck}%
	}
	
	\caption{Structural components of ResNet}
\end{figure*}

Fig.~\ref{fig:basicBlock} shows the basic building block of a ResNet. After every two convolution blocks, there is an input residual mapping added to the output of the blocks which then goes to the ReLu activation layer. The function $F(x)$ is an identity function prevents the residual mapping in adding any additional parameters to the network. Fig.~\ref{fig:bottleNeck} shows a bottleneck block for ResNet 50/101/152 where a three stack layer replaces a two stack layer. The idea for introducing a bottleneck block is to reduce training time constraints for deeper networks, without adding any additional parameters. 

For object re-identification, we used a pre-trained ResNet-50 model on ImagNet dataset.  We also removed the last Fully Connected (FC) layer at the end of the network used for classification and added a 2D average pooling with the kernel size of (16, 8) in order to make the output of the network in the shape of a 1D vector with size of 2048 as an embedded appearance features. In contrast to \cite{ristani2018features}, we did not add any additional FC to prevent increasing the computational complexity at the edge node.

\subsubsection{Mobile Network (Mobilenet)}
Most deep convolution networks have a huge number of parameters and operations making them unsuitable for use in mobile and embedded platforms. Authors in \cite{mobilenet}, developed light-weight deep convolution network which they called MobileNets. They effectively break down a standard convolution into a depthwise and pointwise convolution operation reducing the computational complexity of the net. They also introduce two hyper-parameters, width multiplier and resolution multiplier, which alter the thickness of intermediate layers and resolution of the inputs respectively. They evaluate their model on ImageNet dataset with other state of the art light-weight networks. Following the same trend in \cite{mobileNetV2}, MobileNet-V2 were introduced, which incorporated linear bottleneck layers and inverted residual connections into the previous network reducing the multiply-add operations and number of parameters further but increasing the accuracy.

\begin{figure*}
	\centering
	
	\subfigure[Conventional convolution block]{%
		{\includegraphics[width=0.36\textwidth, trim= 5 10 1 5,clip]{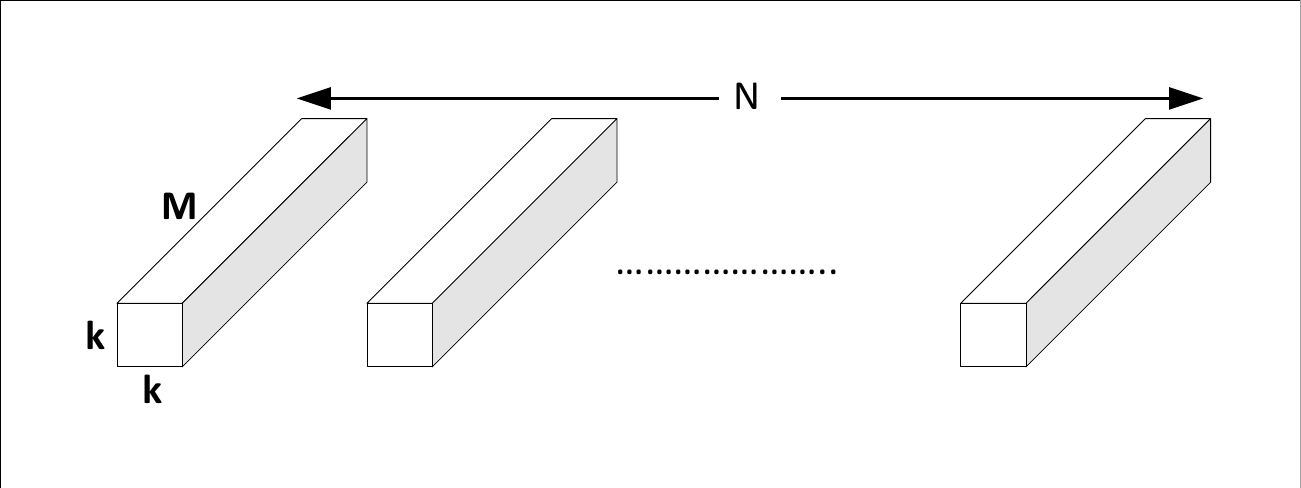}}%
		\label{fig:mna}%
	}\qquad
	\subfigure[Depthwise convolution block]{%
		{\includegraphics[width=0.4\textwidth, trim= 5 10 1 5,clip]{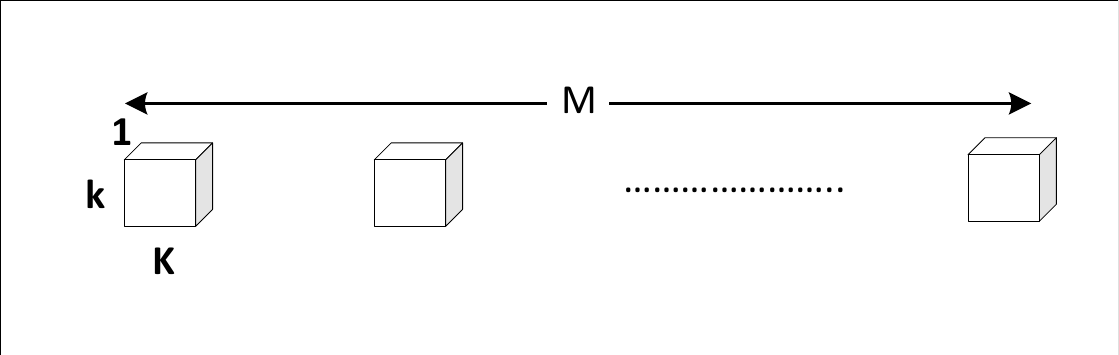}}%
		\label{fig:mnb}%
	}\\
	\subfigure[Pointwise convolution block]{%
	{\includegraphics[width=0.4\textwidth, trim= 5 10 1 5,clip]{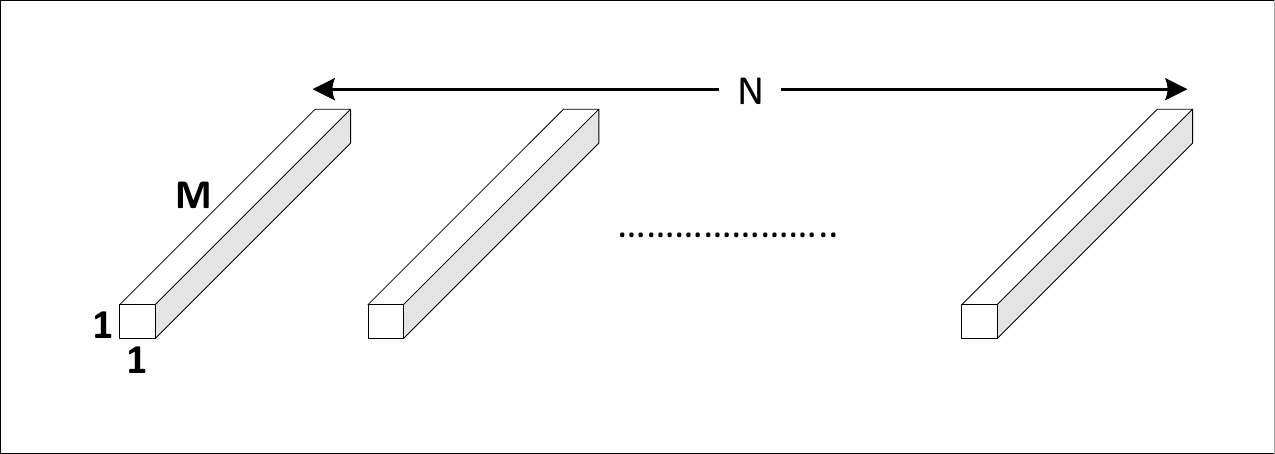}}%
	\label{fig:mnc}%
	}
	\caption{Different \mn~ convolution blocks}
\end{figure*}

In Fig.~\ref{fig:mna}, the normal convolution filters can be seen with the shape of $K \times K \times M \times N$, where $K$ is the size of the filter, $M$ is the input channels, and $N$ is output channels. Andrew et al.[4], transform this convolution into depthwise filtering, Fig.~\ref{fig:mnb}, where each filter is applied to each channel individually and pointwise combination, Fig.~\ref{fig:mnc}, where a $1 \times 1$ filter transforms the filtered features into a new feature map. They show the reduced computation cost and parameters with this approach. MobileNet has 28 layers including depthwise and pointwise layers separately with batch normalization and ReLu activation function. Width multiplier alpha scales the input and output channels by $\alpha \times M$ and $\alpha \times N$. Resolution multiplier does the same thing with the input image resolution hence scaling the computation expense and accuracy trade-off.

We also used a pre-trained MobileNet-V2 model on ImagNet dataset. We removed the classification layer (FC) at the end of the network used for classification and added again a 2D average pooling with the kernel size of (8, 4) in order to make the output of the network in the shape of a 1D vector with size of 1280 as an embedded appearance features.

\subsection{Triplet-loss Function}

Alexander et al. in \cite{Hermans2017InDO}, rekindles the triplet loss network for person re-identification (re-id) with their work. The underlying architecture of a triplet loss network consists of three identical networks which transform the cropped Region of Interest(ROI) into embeddings on a lower dimensional space. One ROI has to be the anchor image, second has to be a positive sample of the anchor and third a negative sample. The basic concept here is to minimize the distance between the anchor and the positive samples and maximize the distance between the anchor and the negative samples in the lower dimensional embedding space. To facilitate such learning, a suitable loss function is used after the embeddings are extracted from the ROIs:

\begin{equation}
Loss = \sum_{i=1}^{n}{\Big[ \alpha + \norm{f^{a}_{i}-f^{p}_{i}}^2 -\norm{f^{a}_{i}-f^{n}_{i}}^2 {\Big]}_{+} },
\end{equation}
where $\alpha$ is margin, $f^{a}$, $f^{p}$, and $f^{n}$ are embedded appearance feature of anchor, positive, and negative samples for the class $i$, respectively. Minimizing $Loss$ function will force all samples of class $i$ to be inside of hypersphere of radius $\alpha$. The dimension of the hypersphere is equal to the size of the output of our networks (2048 for \rn, and 1280 for \mn). Now a drawback here is that the network might only learn easy samples and not hard samples, i.e., hard positives and hard negatives, and be biased towards the easy ones. An example of hard positives is when a person may change his/her clothes and an example of a hard negative is two different persons wearing the same colored clothes/accessories. To overcome this problem, hard sample mining should be accomplished by selecting hard samples for each class after each optimization iteration. In the next iteration, positive and negative samples are selected for class $i$ from the hard samples pool. 
\subsection{Mixed Precision Training}
Since the deep learning approaches are error-tolerant algorithms, designers decrease the accuracy of these networks by lowering the number of bits required to represents weights and biases, and they minimize the introduced error caused by quantization by training the network with reduced precision. However, half precision training needs to overcome two critical challenges of mapping numbers which are too small to be represented in half precision, and vanishing gradient due to limited precision representation. In \cite{micikevicius2018mixed}, they address these problems FP32 master copy of weights, and gradient scaling method during the back-propagation respectively.

For networks used for person identification, we partitioned networks in two single, and half precision categories. We used Apex\footnote{\href{https://github.com/NVIDIA/apex}{https://github.com/NVIDIA/apex}} to assign error-friendly operations, such as convolution and General Matrix Multiply (GeMM) operation, to half precision. During our experiment, we realized if we map the inputs of batch normalization layers in both networks to half precision, training does not converge. We also realized that the loss calculation should also be accomplished in single precision since hard samples are extracted based on loss function and the average distance between anchors and their positives and negatives instances. Lowering the accuracy at loss function will lead to weak hard positives and negatives pool.

\section{Experimental Results}\label{sec:ExpResults}
We evaluate the performance and accuracy of the two \rn~and \mn~networks in this section. We also describe the testing data-sets, the hardware setups, training time, accuracy, throughput, and power consumption for both single  and half precision for each network on three datasets.

\subsection{Learning Parameters and Datasets}
We used DukeMTMC-reID \cite{ristani2016MTMC,zheng2017unlabeled}, CUHK03 \cite{li2014deepreid}, and Market1501 \cite{zheng2015scalable} for evaluating the performance of two netwoks with different training methods. Table \ref{table:training_params} summarizes the hyper parameters of our network. We updated the baseline framework for person re-ID \footnote{\href{https://github.com/huanghoujing/person-reid-triplet-loss-baseline}{https://github.com/huanghoujing/person-reid-triplet-loss-baseline}} in order to support mixed precision training and different network models. We used the combined version of training sets of all three datasets to have better generalization at test phase. We decreased learning rate exponentially after 150 epochs and used Adam optimizer to train both networks.

\begin{table}[h]
	\caption{The training parameters}\label{table:training_params}
	\centering 
	\scalebox{0.8}{
		\begin{tabular}{|c|c|c|c|}
			\hline
			Item & Description \ & Value
			\\[0.1ex]
			\hline                  
			1 &  Batch size  & 128
			\\
			2 &  IDs per batch  & 32
			\\
			3 & Instances per ID & 4
			\\
			4 & Initial learning rate   & $2\times10^{-4}$
			\\
			5  & Input shape (H$\times$W) & (256$\times$128)			
			\\
			6  & Epoch & 300	
			\\
			7  & Margin & 0.3	
			\\
			\hline			
		\end{tabular}
	}
	
\end{table}

\subsection{Accuracy}
\begin{figure}[tbp]
	\centering
	\subfigure[Single Precision]{
		{\begin{tikzpicture}[scale=0.46]
		\begin{axis}[
		ybar,
		bar width=.25cm,
		width=.9\textwidth,
		height=.6\textwidth,
		legend style={
			at={(0.5,-0.2)},
			anchor=north,
			legend columns=-1,
			/tikz/every even column/.append style={column sep=0.1cm}
		},
		enlarge x limits={abs=1.5cm},
		x tick label style={font=\large, rotate=0, anchor=north},
		y tick label style={font=\large},
		symbolic x coords={DukeMTMC,CUHK03,Market1501, Mean},
		xtick=data,
		nodes near coords = \rotatebox{90}{{\pgfmathprintnumber[fixed zerofill, precision=1]{\pgfplotspointmeta}}},
		nodes near coords align={vertical},
		ymin=45,ymax=100,
		ylabel={\%},
		]
		\addplot table[x=interval,y=cmc1]{\singleResNet};
		\addplot table[x=interval,y=cmc5]{\singleResNet};
		\addplot table[x=interval,y=cmc1RR]{\singleResNet};
		\addplot table[x=interval,y=cmc5RR]{\singleResNet};
		\addplot table[x=interval,y=mAP]{\singleResNet};
		\addplot table[x=interval,y=mAPRR]{\singleResNet};
		\legend{CMC-1, CMC-5, CMC-1(R.R), CMC-5(R.R), mAP, mAP(R.R)}
		\end{axis}
		\end{tikzpicture}}
	}\qquad
	\subfigure[Mixed Precision]{
		{\begin{tikzpicture}[scale=0.46]

		\begin{axis}[
		ybar,
		bar width=.25cm,
		width=.9\textwidth,
		height=.6\textwidth,
		legend style={
			at={(0.5,-0.2)},
			anchor=north,
			legend columns=-1,
			/tikz/every even column/.append style={column sep=0.1cm}
		},
		enlarge x limits={abs=1.5cm},
		x tick label style={font=\large, rotate=0, anchor=north},
		y tick label style={font=\large},
		symbolic x coords={DukeMTMC,CUHK03,Market1501, Mean},
		xtick=data,
		nodes near coords = \rotatebox{90}{{\pgfmathprintnumber[fixed zerofill, precision=1]{\pgfplotspointmeta}}},
		nodes near coords align={vertical},
		ymin=50,ymax=100,
		ylabel={\%},
		]
		\addplot table[x=interval,y=cmc1]{\halfResNet};
		\addplot table[x=interval,y=cmc5]{\halfResNet};
		\addplot table[x=interval,y=cmc1RR]{\halfResNet};
		\addplot table[x=interval,y=cmc5RR]{\halfResNet};
		\addplot table[x=interval,y=mAP]{\halfResNet};
		\addplot table[x=interval,y=mAPRR]{\halfResNet};
		\legend{CMC-1, CMC-5, CMC-1(R.R), CMC-5(R.R), mAP, mAP(R.R)}
		\end{axis}
		\end{tikzpicture}}
	}
	
	\caption{ResNet-50 accuracy evaluation on three different benchmarks. We trained the model for two different precision configuration, one single precision (a) and mixed precision (b).} \label{fig:accResnet}
\end{figure}
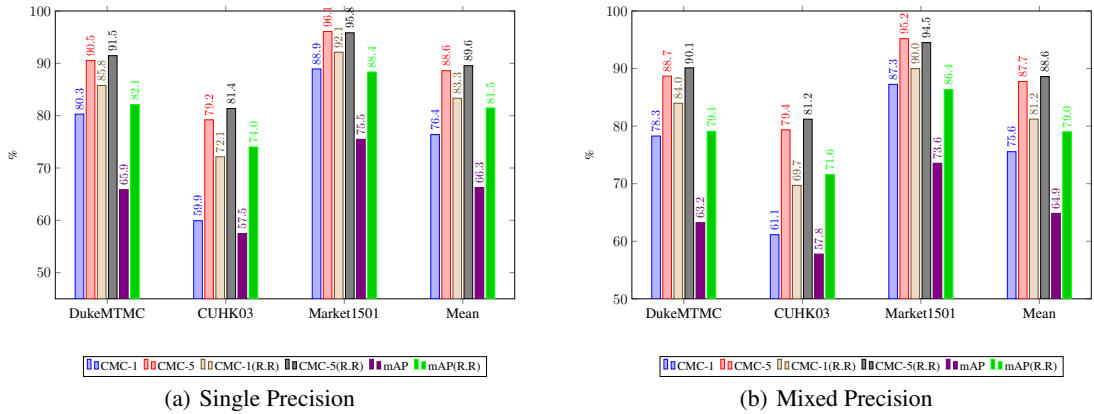

Fig.~\ref{fig:accResnet} compares the accuracy results for baseline and \rn~half precision. The Re-Ranking (RR) \cite{zhong2017re} method can improve the mAP on the average of 12\% for both half and single precision. The CUHK03 benefits the highest improvement by applying the RR method among other datasets. Based on the results, we can realize that half precision only degrades 0.9\% on the average concerning single precision for all three CMC-(1, 5) and mAP. 

\begin{figure}[tbp]
	\centering
	\subfigure[Single Precision]{
		{\begin{tikzpicture}[scale=0.45]
			\begin{axis}[
			ybar,
			bar width=.25cm,
			width=.9\textwidth,
			height=.6\textwidth,
			legend style={
				at={(0.5,-0.2)},
				anchor=north,
				legend columns=-1,
				/tikz/every even column/.append style={column sep=0.1cm}
			},
			enlarge x limits={abs=1.5cm},
			x tick label style={font=\large, rotate=0, anchor=north},
			y tick label style={font=\large},
			symbolic x coords={DukeMTMC,CUHK03,Market1501, Mean},
		xtick=data,
		nodes near coords = \rotatebox{90}{{\pgfmathprintnumber[fixed zerofill, precision=1]{\pgfplotspointmeta}}},
		nodes near coords align={vertical},
		ymin=45,ymax=100,
		ylabel={\%},
		]
		\addplot table[x=interval,y=cmc1]{\singleMobileNet};
		\addplot table[x=interval,y=cmc5]{\singleMobileNet};
		\addplot table[x=interval,y=cmc1RR]{\singleMobileNet};
		\addplot table[x=interval,y=cmc5RR]{\singleMobileNet};
		\addplot table[x=interval,y=mAP]{\singleMobileNet};
		\addplot table[x=interval,y=mAPRR]{\singleMobileNet};
		\legend{CMC-1, CMC-5, CMC-1(R.R), CMC-5(R.R), mAP, mAP(R.R)}
		\end{axis}
		\end{tikzpicture}}
	}\qquad
	\subfigure[Mixed Precision]{
		{\begin{tikzpicture}[scale=0.45]
			\begin{axis}[
			ybar,
			bar width=.25cm,
			width=.9\textwidth,
			height=.6\textwidth,
			legend style={
				at={(0.5,-0.2)},
				anchor=north,
				legend columns=-1,
				/tikz/every even column/.append style={column sep=0.1cm}
			},
			enlarge x limits={abs=1.5cm},
			x tick label style={font=\large, rotate=0, anchor=north},
			y tick label style={font=\large},
			symbolic x coords={DukeMTMC,CUHK03,Market1501, Mean},
		xtick=data,
		nodes near coords = \rotatebox{90}{{\pgfmathprintnumber[fixed zerofill, precision=1]{\pgfplotspointmeta}}},
		nodes near coords align={vertical},
		ymin=45,ymax=100,
		ylabel={\%},
		]
		\addplot table[x=interval,y=cmc1]{\halfMobileNet};
		\addplot table[x=interval,y=cmc5]{\halfMobileNet};
		\addplot table[x=interval,y=cmc1RR]{\halfMobileNet};
		\addplot table[x=interval,y=cmc5RR]{\halfMobileNet};
		\addplot table[x=interval,y=mAP]{\halfMobileNet};
		\addplot table[x=interval,y=mAPRR]{\halfMobileNet};
		\legend{CMC-1, CMC-5, CMC-1(R.R), CMC-5(R.R), mAP, mAP(R.R)}
		\end{axis}
		\end{tikzpicture}
	}
	}
	
	\caption{MobileNetV2 accuracy evaluation on three different benchmarks. We trained the model for two different precision configuration, one single precision (a) and mixed precision (b).} \label{fig:mnAcc}
\end{figure}
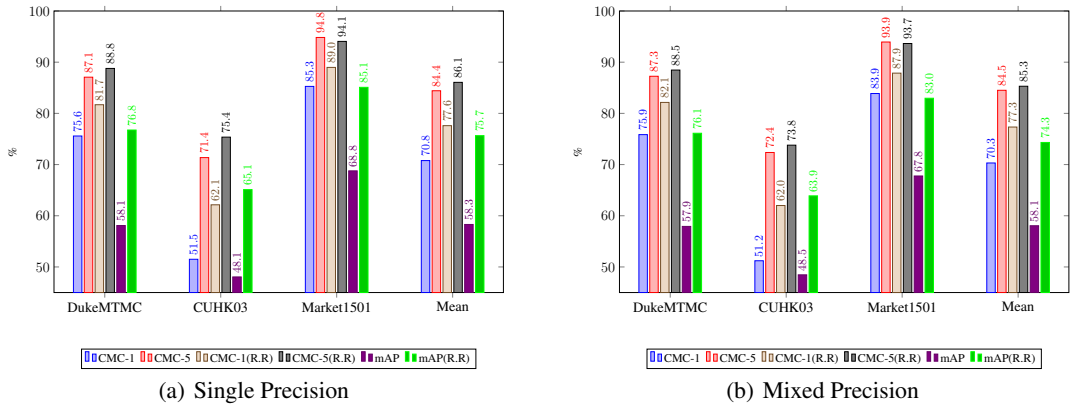

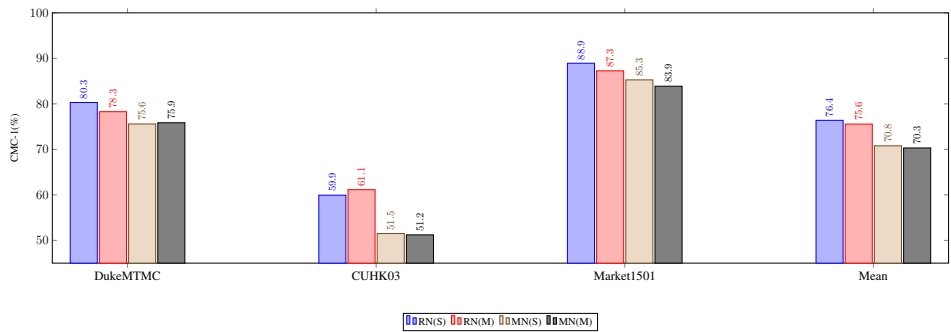
\begin{figure}[tbp]
	\centering
	\begin{tikzpicture}[scale=0.4]
	\begin{axis}[
	ybar,
	bar width=.9cm,
	width=1.9\textwidth,
	height=.6\textwidth,
	legend style={
		at={(0.5,-0.2)},
		anchor=north,
		legend columns=-1,
		/tikz/every even column/.append style={column sep=0.1cm}
	},
	x tick label style={font=\large, rotate=0, anchor=north},
	y tick label style={font=\large},			
	enlarge x limits={abs=2.5cm},
	symbolic x coords={DukeMTMC,CUHK03,Market1501,Mean},
	xtick=data,
	nodes near coords = \rotatebox{90}{{\pgfmathprintnumber[fixed zerofill, precision=1]{\pgfplotspointmeta}}},
	nodes near coords align={vertical},
	ymin=45,ymax=100,
	ylabel={CMC-1(\%)},
	]
	\addplot table[x=interval,y=ResNet50-S]{\allNets};
	\addplot table[x=interval,y=ResNet50-M]{\allNets};
	\addplot table[x=interval,y=MobileNetV2-S]{\allNets};
	\addplot table[x=interval,y=MobileNetV2-M]{\allNets};
	legend style={nodes={scale=1.5, transform shape}, font=\large}
	\legend{RN(S), RN(M), MN(S), MN(M)}
	\end{axis}
	\end{tikzpicture}
	
	\caption{A CMC-1 Comparison of two networks ResNet50 (RN) and MobileNetV2 (MN) for three different training and inference approaches. } \label{fig:RnMn}
\end{figure}

\begin{figure}[htbp]
	\centering
	\subfigure[\rn: Single precision (left), half precision (right)]{
		\includegraphics[width=0.77\textwidth, trim= 5 1 1 5,clip]{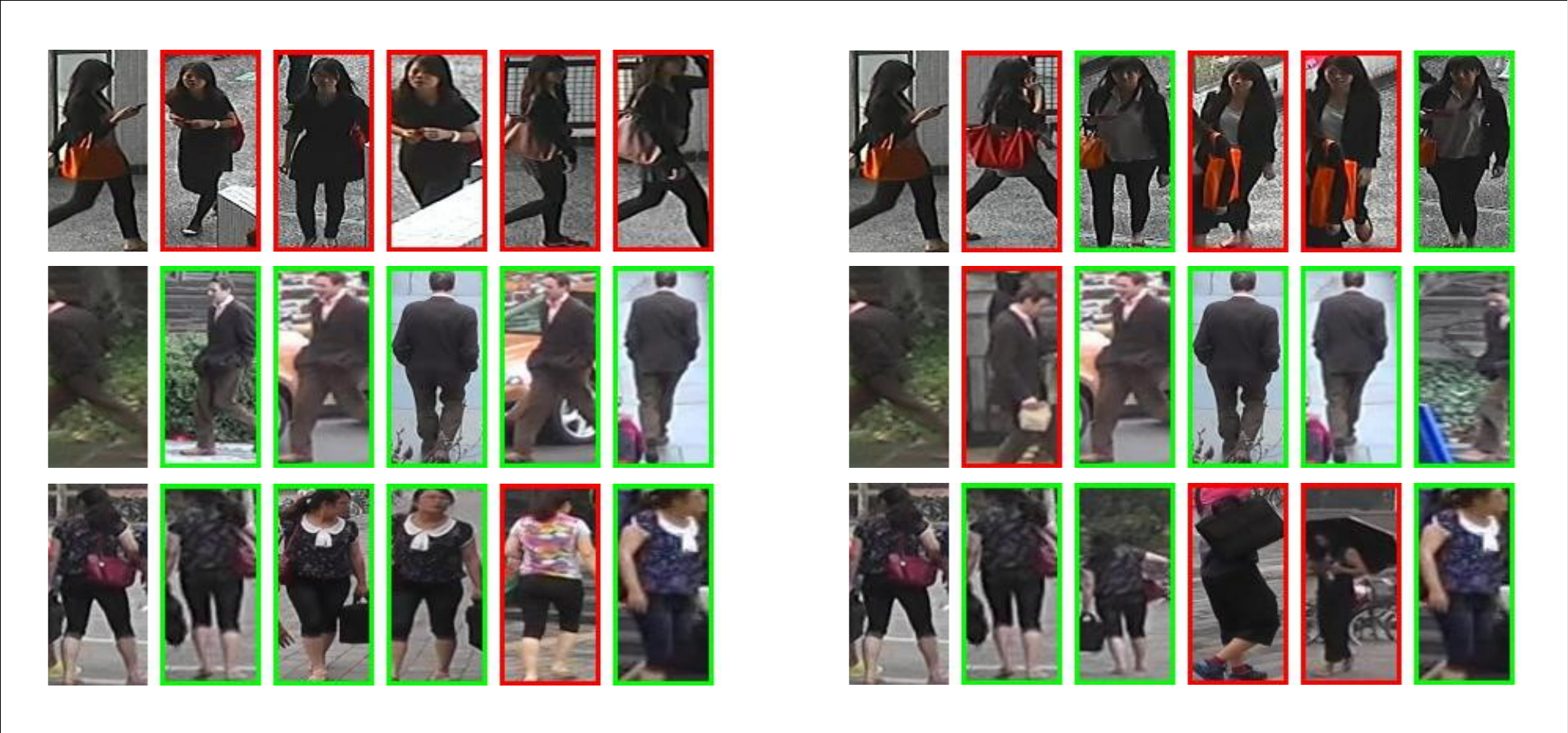}
 	}
	\subfigure[\mn: Single precision (left), half precision (right)]{
		\includegraphics[width=0.77\textwidth, trim= 5 10 1 5,clip]{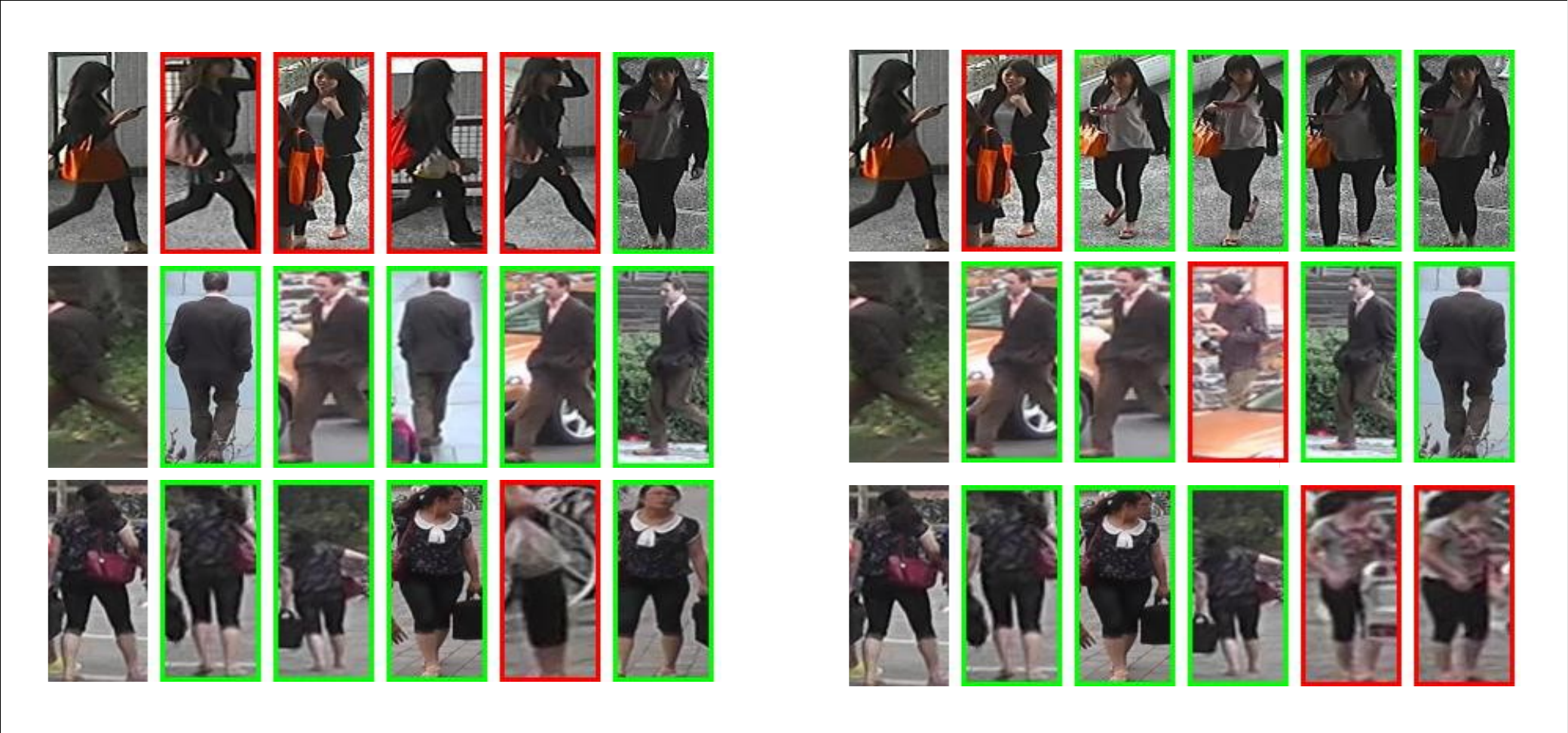}
		
	}
	\caption{A qualitative comparison of two networks with different precision. The images without the bounding box in each sample are queries, and five images in front of it are the first five ranked samples in the gallery. We marked true and false detected with green and red boxes respectively.} \label{fig:qaulComp}
\end{figure}

Fig.~\ref{fig:mnAcc} depicts the \mn~ model performance for both single and half precision in a similar approach. As we can see single precision negligibility deteriorate the CMC-1 performance for 0.5\%. Based on side by side CMC-1 comparison on Fig.~\ref{fig:RnMn} for both \rn~and \mn~ network, we can realize that the \mn~half precision is negligibly 5.6\% less than baseline. We also compare the results qualitatively in Fig.~\ref{fig:qaulComp}. We selected randomly three queries from each dataset and sorted the nearest objects based on Euclidean distance from the gallery considering the embedded appearance feature extracted without applying the RR method.

\pgfplotstableread[row sep=\\,col sep=&, header=true]{
	interval      & cmc1   & cmc5   & cmc1RR  & cmc5RR  & mAP    & mAPRR \\
	DukeMTMC      & 80.30  & 90.53  & 85.77   & 91.47   & 65.86  & 82.11 \\
	CUHK03        & 59.93  & 79.21  & 72.14   & 81.36   & 57.47  & 74.01 \\
	Market1501    & 88.93  & 96.08  & 92.13   & 95.84   & 75.48  & 88.35 \\
	Mean          & 76.38  & 88.60  &  83.34  &  89.55  &  66.27 & 81.47 \\
}\singleResNet

\pgfplotstableread[row sep=\\,col sep=&, header=true]{
	interval     & cmc1   & cmc5   & cmc1RR  & cmc5RR  & mAP    & mAPRR \\
	Market1501   & 87.26  & 95.19  & 89.99   & 94.51   & 73.55  & 86.36 \\
	CUHK03       & 61.14  & 79.36  & 69.71   & 81.21   & 57.77  & 71.58 \\
	DukeMTMC     & 78.28  & 88.69  & 83.98   & 90.13   & 63.23  & 79.09 \\
	Mean         & 75.56  & 87.74  & 81.22   & 88.61   & 64.85  & 79.01 \\
}\halfResNet

\pgfplotstableread[row sep=\\,col sep=&, header=true]{
	interval      & cmc1   & cmc5   & cmc1RR  & cmc5RR  & mAP    & mAPRR \\
	Market1501    & 85.27  & 94.83  & 88.98   & 94.06   & 68.77  & 85.09 \\
	CUHK03        & 51.50  & 71.36  & 62.14   & 75.36   & 48.05  & 65.11 \\
	DukeMTMC      & 75.58  & 87.07  & 81.69   & 88.78   & 58.09  & 76.75 \\
	Mean 		  & 70.78  & 84.42  & 77.60   & 86.06   & 58.30  & 75.65 \\
}\singleMobileNet

\pgfplotstableread[row sep=\\,col sep=&, header=true]{
	interval     & cmc1   & cmc5   & cmc1RR  & cmc5RR  & mAP    & mAPRR \\
	Market1501   & 83.88  & 93.94  & 87.86   & 93.65   & 67.77  & 82.96 \\
	CUHK03       & 51.21  & 72.36  & 62.00   & 73.79   & 48.46  & 63.89 \\
	DukeMTMC     & 75.85  & 87.25  & 82.14   & 88.46   & 57.92  & 76.12 \\
	Mean         & 70.31  & 84.51  & 77.33   & 85.30   & 58.05  & 74.32 \\
}\halfMobileNet

\pgfplotstableread[row sep=\\,col sep=&, header=true]{
	interval     & ResNet50-S   & ResNet50-M  & MobileNetV2-S  & MobileNetV2-M \\
	Market1501   & 88.93        & 87.26       & 85.27          & 83.88 \\
	CUHK03       & 59.93        & 61.14       & 51.50          & 51.21 \\
	DukeMTMC     & 80.30        & 78.28       & 75.58          & 75.85 \\
	Mean         & 76.38        & 75.56       & 70.78          &70.31  \\
}\allNets

\subsection{Training time}
We used the system described in Table \ref{table:system} to train two \rn~and \mn~networks. Table \ref{table:training_time} shows the results of training time on the server for different system configuration. As we can observe, half precision can improve training time to $1.20 \times$  on the average for both networks, and \mn~half precision can upgrade it to $1.75 \times$ with respect to the baseline model.

\begin{table}[!htbp]
	\caption{Training system configuration}
	\centering 
	\scalebox{0.8}{
		\begin{tabular}{|c|c|c|}
			\hline
			Item & Description \ & Value
			\\[0.1ex]
			\hline  
			1 & Processor Info & Intel(R) Xeon(R) CPU E5-2640                
			\\
			2 &  CPU Cores  & 40
			\\
			3 &  GPUs & 2 $\times{}$ nVidia TITAN V
			\\
			4 &  OS Version & Ubuntu 18.04 LTS
			\\
			5 & Memory & 96GB/94.2GB Available
			\\
			\hline			
		\end{tabular}
		\label{table:system}
	}
\end{table}

\begin{table}[tbp]
	\caption{Training elapsed time of two networks with different precision}\label{table:training_time}
	\centering 
	\scalebox{0.8}{
		\begin{tabular}{|c|c|c|c|}
			\hline
			Item & Description \ & Value (mins.)
			\\[0.1ex]
			\hline                  
			1 &  \rn~(single)  & 242.65
			\\
			2 &  \rn~(half)  & 174.45
			\\
			3 & \mn~(single) & 140.3
			\\
			4 & \mn~(half)   & 138.1
			\\			
			\hline			
		\end{tabular}
	}
	
\end{table}

\subsection{Edge Node Evaluation}
We evaluated system performance in respect to power consumption and inference time on Nvidia Xavier embedded node. Table \ref{table:systemInfr} summarizes the hardware resources on this board. We extracted the Open Neural Network eXchange (ONNX) format representation of all four network configurations and uploaded them on the edge side. As it is depicted in Table \ref{table:modelsize}, we can reach to 18.92$\times$ model size compression ratio over the baseline model for \mn~half precision. We mapped the half-precision types of both networks on Deep Learning Accelerators (DLA) and single precision to GPU Volta cores. We also set the Xavier power mode to MAX-N\footnote{\href{https://developer.nvidia.com/embedded/jetson-agx-xavier-dl-inference-benchmarks}{https://developer.nvidia.com/embedded/jetson-agx-xavier-dl-inference-benchmarks}}. 
\begin{table}[tbp]
	\caption{The edge node hardware configuration}
	\centering 
	\scalebox{0.8}{
		\begin{tabular}{|c|c|c|}
			\hline
			Item & Description \ & Value
			\\[0.1ex]
			\hline  
			1 & Processor Info & ARM v8.2 64-bit CPU                
			\\
			2 &  CPU Cores  & 8
			\\
			3 &  GPUs &  512-Core Volta GPU
			\\
			4 & DL Accelerators & 2
			\\			
			5 &  OS Version & Ubuntu 18.04 LTS
			\\
			6 & Memory & 16GB/15.4GB Available
			\\
			\hline			
		\end{tabular}
		\label{table:systemInfr}
	}
\end{table}

The inference time (Table \ref{table:infrTime}) and power consumption (Table \ref{table:power}) is acquired for the batch size of 16. We obtained both timing performance and power consumption only for extracting features, and we did not consider model loading and other pre-processing tasks. \mn~ half precision improves the inference throughput 3.25$\times$ and reaches to 27.77fps, while it only consumes 6.48W. As the hardware warm-up was same for both \mn~half-precision and single precision, we did not observe any power consumption improvement for this network. 

\begin{table}[!h]
	\caption{Model sizes of two networks.}
	\centering 
	\scalebox{0.7}{
		\begin{tabular}{|c|c|c|c|c|}
			\hline
			\multirow{2}{*}{Network} & \multicolumn{2}{c|}{Model Size (MB)} & \multicolumn{2}{c|}{Improvement($\times$)}
			\\\cline{2-5}
			& Single Precision & Mixed Precision & Per Same Model & Over the Baseline
			\\[0.1ex]
			\hline                  
			ResNet-50 &  94.6  & 47.7 & 1.98 & \multirow{2}{*}{18.92}
			\\
			MobileNetV2 &  9.4 & 5.0 & 1.88 &
			\\
			\hline			
		\end{tabular}
		\label{table:modelsize}
	}
\end{table}
\begin{table}[!h]
	\caption{Throughput performance on Nvidia Xavier.}
	\centering 
	\scalebox{0.7}{
		\begin{tabular}{|c|c|c|c|c|}
			\hline
			\multirow{2}{*}{Network} & \multicolumn{2}{c|}{Throughput (fps)} & \multicolumn{2}{c|}{Improvement ($\times$)}
			\\\cline{2-5}
		& Single Precision & Mixed Precision & Per Same Model & Over the Baseline 
			\\[0.1ex]
			\hline                  
			ResNet-50 &  8.54  & 21.71 & 2.54 & \multirow{2}{*}{3.25}
			\\
			MobileNetV2 & 20 & 27.77 & 1.38 &
			\\
			\hline			
		\end{tabular}
		\label{table:infrTime}
	}
\end{table}
\begin{table}[!h]
	\caption{Power consumption on Nvidia Xavier.}
	\centering 
	\scalebox{0.7}{
		\begin{tabular}{|c|c|c|c|c|}
			\hline
			\multirow{2}{*}{Network} & \multicolumn{2}{c|}{Power (W)} & \multicolumn{2}{c|}{Improvement ($\times$)}
			\\\cline{2-5}
			& Single Precision & Mixed Precision & Per Same Model & Over the Baseline
			\\[0.1ex]
			\hline                  
			ResNet-50 &  9.45  & 7.86 & 1.2 & \multirow{2}{*}{1.45}
			\\
			MobileNetV2 &  6.48 & 6.48 & 1 &
			\\
			\hline			
		\end{tabular}
		\label{table:power}
	}
\end{table}

\section{Conclusion}\label{sec:Conclusion}
In this paper, we present a light-weight person re-identification method based on \mn. We even improved the performance of the edge node to the next level by mapping models to half precision. The experimental results elucidate that mixed precision training can achieve real-time re-identifying persons at the frame rate of 27.77 per second by only consuming 6.48W. Our finding of network partitioning for mixed precision training is orthogonal to other person re-identification based on deep learning paradigm and can be applied to improve the overall system performance.


\bibliographystyle{splncs04}
\scriptsize
\bibliography{paper}

\end{document}